\documentclass[conference]{IEEEtran}
\IEEEoverridecommandlockouts
\usepackage{cite}
\usepackage{amsmath,amssymb,amsfonts}
\usepackage{algorithm}
\usepackage{algorithmicx}
\usepackage{algpseudocode}
\usepackage{graphicx}
\usepackage{textcomp}
\usepackage{xcolor}
\usepackage{bm}

\def\BibTeX{{\rm B\kern-.05em{\sc i\kern-.025em b}\kern-.08em
    T\kern-.1667em\lower.7ex\hbox{E}\kern-.125emX}}
\begin{document}

\title{Evolution Attack On Neural Networks
%\thanks{YiGui Luo is the corresponding author(e-mail: luoyigui@tongji.edu.cn).}
}

\author{\IEEEauthorblockN{YiGui Luo, RuiJia Yang, Wei Sha, WeiYi Ding, YouTeng Sun, YiSi Wang}
\IEEEauthorblockA{\textit{School of Software Engineering and} \\
\textit{Institute of Intelligent Vehicles, TongJi University}\\
ShangHai, China \\
luoyigui@tongji.edu.cn, 1641491@tongji.edu.cn}
}

\maketitle

\begin{abstract}
Many studies have been done to prove the vulnerability of neural networks to adversarial example. A trained and well-behaved model can be fooled by a visually imperceptible perturbation, i.e., an originally correctly classified image could be misclassified after a slight perturbation. In this paper, we propose a black-box strategy to attack such networks using an evolution algorithm. First, we formalize the generation of an adversarial example into the optimization problem of perturbations that represent the noise added to an original image at each pixel. To solve this optimization problem in a black-box way, we find that an evolution algorithm perfectly meets our requirement since it can work without any gradient information. Therefore, we test various evolution algorithms, including a simple genetic algorithm, a parameter-exploring policy gradient, an OpenAI evolution strategy, and a covariance matrix adaptive evolution strategy. Experimental results show that a covariance matrix adaptive evolution Strategy performs best in this optimization problem. Additionally, we also perform several experiments to explore the effect of different regularizations on improving the quality of an adversarial example.
\end{abstract}


\section{Introduction}
Currently, deep convolutional neural networks (DCNNs) have become the state-of-art learning model in the computer vision field. Since AlexNet\cite{b1} won the ImageNet Large Scale Visual Recognition Competition (ILSVRC) in 2012, an increasing number of researchers have been attracted to DCNNs due to their powerful abilities of learning and feature extraction. Up to now, almost every advanced solution for image problems are based on DCNNs, such as image classification, target detection\cite{b2}\cite{b3}, semantic segmentation\cite{b4}\cite{b5}, and pose estimation\cite{b6}. With the development of computer vision technology, many intelligent applications have emerged, e.g., face recognition for identity authentication and visual perception for intelligent vehicles. Therefore, the security of these intelligent systems has become increasingly more important.
\begin{figure}[tbp]
\centerline{\includegraphics[scale=0.75]{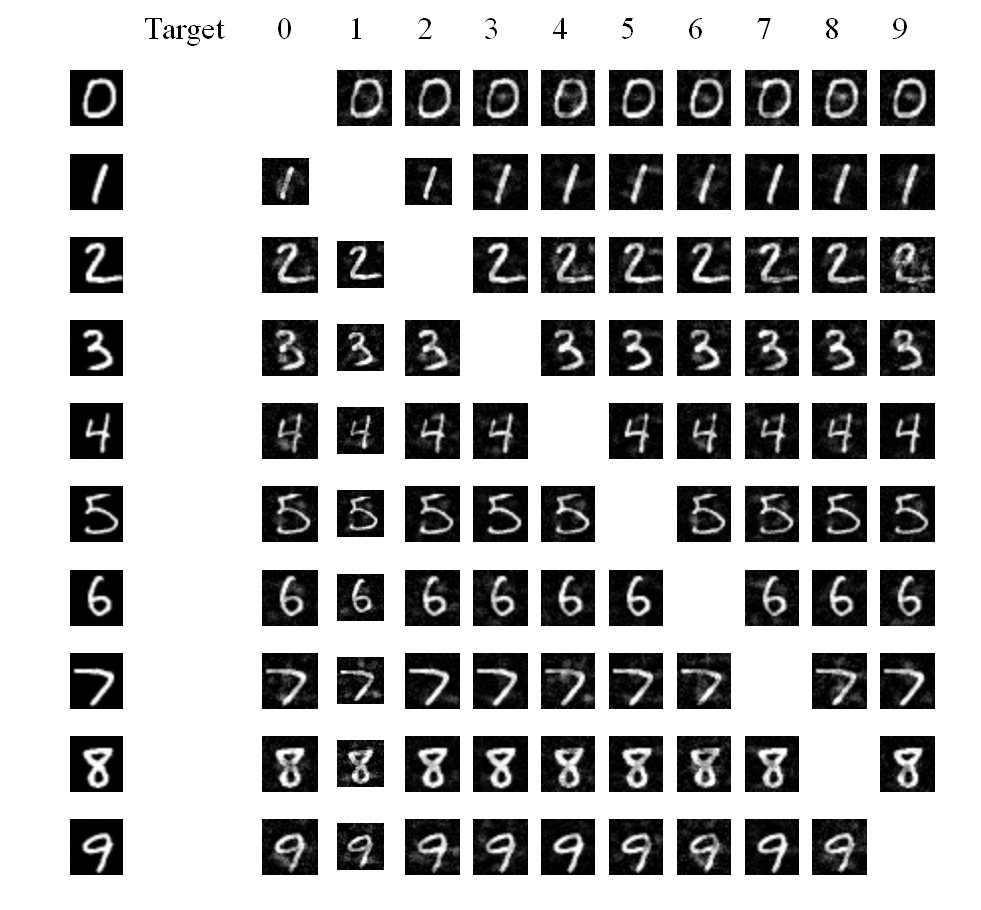}}
\caption{Adversarial example generated by evolution attack. Left, original clean image; right, adversarial example for different misclassification targets.}
\label{fig}
\end{figure}

Although DCNNs have shown impressive results in such fields, more attention must be paid to their weaknesses, namely, that they do not constitute a robust model and are vulnerable to adversarial example attacks\cite{b8}. For basic image classification problems in computer vision, assuming a trained classifier exists, the purpose of such attacks is to find the perturbations that can make the input image able to fool the classifier. After being perturbed, the image that is originally correctly classified will be misclassified by the classifier. The perturbed image is called an adversarial example. In addition, the perturbations must be as small as possible; otherwise, the abnormal image can be easily found and rejected manually.

The issue in generating a qualified, quality adversarial example is to find the perturbations that meet the following requirements:
\begin{itemize}
\item Perturbations that lead to the misclassification of an image that is correctly classified.
\item Perturbations need to be as unperceptive to people as much as possible.

\end{itemize}

Based on access right to model gradient information, two kinds of attacks exist: white-box attacks and black-box attacks.

A white-box attack is one in which attackers need to know detailed information about the target DCNN model. For example, the efficient attack algorithms fast gradient sign (FGS)\cite{b9} and fast gradient value (FGV)\cite{b7} require that the gradient information of classifier output for every input pixel element be known. It is implied that these methods will not work without knowledge of the internal architecture of the classifier. Considering an actual situation, it is difficult for an attacker to know such information, so the black-box attack method is much more dangerous than the white-box attack method.

In contrast to a white-box attack, the basic and important requirement for a black-box attack is to generate an adversarial example without detailed information of the classifier system. The only knowledge required is the output vector of the classifier.

To attack a target model without gradient information, we find that the intelligent optimization method called an evolution algorithm (EA)\cite{b21}, whose primary characteristic, i.e.,  no demand for gradient information, perfectly meets the requirements of a black-box attack if the problem of adversarial example generation can be transformed into the optimization problem of perturbation.
EAs comprise another large field of study, compared to numerical analysis optimization methods like gradient descent and the Newton method, and are closer to space searching. To solve the above optimization problem, we propose an advanced EA named the covariance matrix adaptation evolution strategy (CMA-ES)\cite{b20}\cite{b22}\cite{b23}. In addition, we test other EAs for comparison, such as a simple genetic algorithms (GA)\cite{b16}, a parameter-exploring policy gradient (PEPG)\cite{b17}\cite{b18}, and the OpenAI evolution strategy (OpenAI ES)\cite{b19}. To improve the quality of adversarial examples, we also performed several experiments to explore the effects using different regularizations.

The main work presented in this paper comprises the following:
\begin{itemize}
%\item Formalizing the transformation of process of generating an adversarial example into an optimization problem.
\item Formalizing the process of generating an adversarial example.
\item Successfully generating the adversarial example by our black-box attack strategy. As the most indispensable part, the efficiency of the EA directly determines the quality of the adversarial example. Therefore, we test different EAs, such as a GA, PEPG, OpenAI ES, and CMA-ES in our strategy.
\item Considering that our method can generate an adversarial example of 100\%, the quality, i.e., the similarity between the original and perturbed images, is the most valuable evaluation index for our attack method. Thus, we evaluate the adversarial examples generated by different EAs. In addition, we also try different regularizations of our optimization target to reduce the perturbations.
\end{itemize}

\section{Related Work}
Although DCNNs have been proven to be a powerful learning model, Szegedy et al.\cite{b8} first found that a hardly perceptible perturbation would lead to misclassification of the trained Network. Goodfellow et al. \cite{b9} applied the FGS algorithm to generate an adversarial example and explained that the linear nature of DCNNs was responsible for its vulnerability:

\begin{equation}
\bm{\eta}=\epsilon sign(\nabla_{\bm x} J(\bm \theta,\bm x,y) ).
\end{equation}

Rozsa et al.\cite{b7} improved the FGS and proposed the FGV algorithm:
\begin{equation}
\bm{\eta}=\epsilon (\nabla_{\bm x} J(\bm \theta,\bm x,y) ).
\end{equation}
Moosavi-Dezfooli et al. \cite{b10} suggested an iteration method named DeepFool to minimize the perturbation $\bm \eta$, which is sufficient to change the predicted result. The existence of universal perturbations was shown in \cite{b12}. Baluja et al.\cite{b14} proposed the adversarial transformation network, which is similar to an autoencoder, to generate an adversarial example. Xie et al.\cite{b13} developed an adversarial example for semantic segmentation and object detection.

The above works all concern a white-box attack. Considering a black-box attack is more dangerous in the real world, Papernot et al.\cite{b11} proposed the black-box attack strategy, in which the substitute model needs to be trained using the outputs of the original model as labels. Thus, an adversarial example based on detailed information of the new substitute model can also fool the original model.

For adversarial defense, Goodfellow et al.\cite{b9} found that adversarial training could reduce network overfitting and, in turn, reinforce the robustness. Papernot et al.\cite{b15} applied defensive distillation to improve the robustness of the network. Lee et al.\cite{b24} tried to use a generative adversarial network (GAN) to train a robust classifier. Its generator is supposed to generate perturbations and its classifier is supposed to correctly classify both the original and perturbed images.

\section{CMA-ES}

\begin{figure*}[tbp]\label{fig1}
\centerline{\includegraphics[scale=0.8]{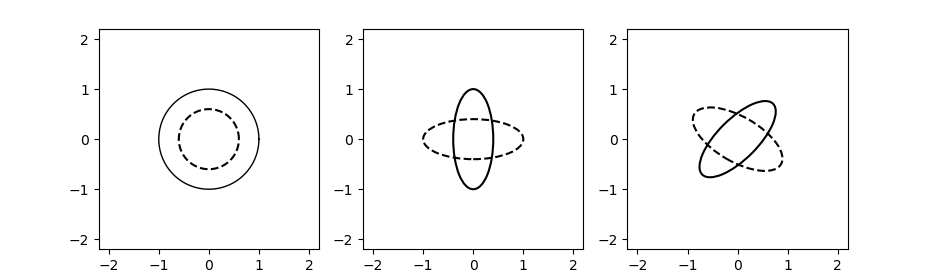}}
\caption{Ellipsoids depicting one-$\sigma$ lines of equal density of six different normal distributions. Left:$\mathcal N(\bm 0,\sigma\bm I)$. Middle: $\mathcal N(\bm 0,\bm D^2)$. Right:$\mathcal N(\bm 0,\bm C)$. $\bm D$ is a diagonal matrix. $\bm C$ is a positive-definite covariance matrix}
\end{figure*}

Below, we introduce the CMA-ES, which, as one of the most advanced EAs, plays the most significant role in the black-box attack method proposed in this paper.

Before the official introduction of CMA-ES, we briefly discuss the basic characteristics and history of EAs.
Different from the numerical analysis optimization methods, such as the gradient descent or Newton methods, an EA requires fewer assumptions and can work without the gradient information. Therefore, it can be seen as the black-box optimization method.

In the beginning, EA was developed from the central idea of Darwin's theory of evolution, i.e., survival of the fittest. Assuming there are large numbers of populations, after natural selection, the survival populations, called elite populations, recombine with each other to generate the new generation of populations that are supposed to have better survival ability then the previous generation. In an EA, the survival ability can be described as a fitness function, i.e., one that is equal to the loss function in an optimization problem.
The above introduction can be seen as the basic EA process. With the passage of time, increasingly more innovations have occurred in EAs, including various methods of population generation, selection, and recombination.

Simple EAs behave well in low-dimensional optimization problems, but nevertheless show poor ability when facing high-dimensional optimization problems. Compared to gradient descent, an EA is more like the space searching method, which means it must face the "dimension explosion" problem with the increase in problem dimensionality.

We submit that CMA-ES can improve the optimization efficiency to some extent. Compared to other EAs, the largest difference in CMA-ES is that it introduces a priori knowledge of problem distribution. Unlike GA and other simple EAs that treat different dimensions independently, CMA-ES regards the optimization problem as a multivariate normal distribution defined by the mean vector and covariance matrix. Every generation of populations is sampled from the current multivariate normal distribution, further, selection and recombination are designed to update the mean vector and covariance matrix to make the distribution closer to the local optimal point. The multivariate normal distribution $\mathcal{N}(\bm m, \bm C)$ is written in \eqref{easy sample}:

\begin{equation}\label{easy sample}
	\begin{split}
	\mathcal{N}(\bm m, \bm C) &\sim \bm m + \mathcal N (\bm 0,\bm C) \\
	&\sim \bm m + \bm C ^{1/2}\mathcal N (\bm 0,\bm I)\\
    &\sim \bm m + \bm B \bm D \underbrace {\bm B^{\mathsf T} \mathcal N (\bm 0,\bm I)}_{\sim \mathcal N(\bm 0,\bm I)} \\
    &\sim \bm m + \bm B \underbrace {\bm D \mathcal N (\bm 0,\bm I)}_{\sim \mathcal N(\bm 0,\bm D^2)}.
	\end{split}
\end{equation}

\begin{figure*}[htbp]
\centerline{\includegraphics[scale=0.8]{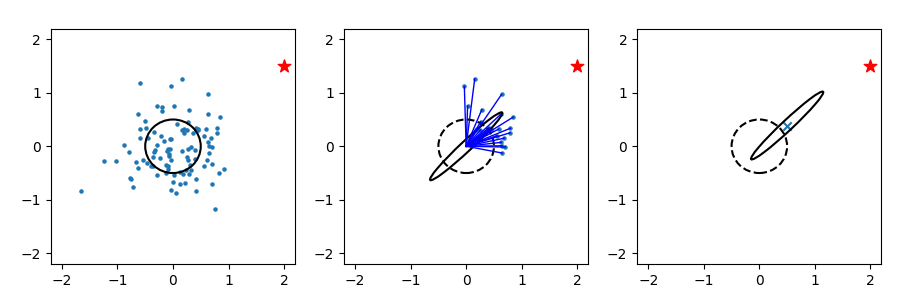}}
\caption{Explanation for CMA-ES evolution process, which is designed to find the minimum point of $(x_1-2)^2+(x_2-1.5)^2$.
Left: Sampling the populations from the current multivariate normal distribution. Middle: Selecting the elite populations and updating the covariance matrix. Right: Moving the mean vector.}
\label{fig_3}
\end{figure*}
The covariance matrix $\bm C\in\mathbb{R}^{n*n}$ is symmetric and positive definite, so $\bm C= \bm B \bm D^2 \bm B^{\mathsf T}$ is the eigen-decomposition of $\bm C$, and then $\bm C^{1/2}= \bm B \bm D \bm B^{\mathsf T}$.

$\mathcal N (\bm 0,\bm I)$ produces a spherical(isotropic) distribution, as in the left-hand panel of Fig.~\ref{fig1}. $\bm D$ scales the spherical distribution within the coordinate axes as in the middle panel of Fig.~\ref{fig1}. $\bm D\mathcal N (\bm 0,\bm I)\sim\mathcal N (\bm 0,\bm D^2)$ has n independent components. The matrix $\bm D$ can be interpreted as an individual step-size matrix and its diagonal entries are the standard deviations of the components. $\bm B$ defines a new orientation for the ellipsoid, where the new principal axes of the ellipsoid correspond to the columns of $\bm B$.

The transformation in \eqref{easy sample} helps sample the population from the current distribution in a computer considering that $\mathcal N(\bm 0,\bm I)$ actually represents the independent normal distribution in every dimension.

Every element in the covariance matrix is the covariance between every two dimensions. Considering the real complex optimization problem is always highly coupled in different dimensions, and the covariance matrix can represent such a complex relationship in some ways if the iteration update is effective. The effective measure of the update will increase the variance along the efficient searching directions. In fact, thinking about the coordinate transformation of a vector in a new coordinate system in principal component analysis (PCA), the main directions represented in the covariance matrix are equivalent to the base vector in the new coordinate system.

The main selection and recombination formed in the CMA-ES include two kinds of methods to adapt the covariance matrix and one kind of method to update the mean vector after selection.

\subsection{Rank-1 Update}
Roughly speaking, a rank-1 update is a historical accumulation of the trajectory of the elite populations' mean vector $\bm m\in\mathbb{R}^n$. Thus, CMA-ES defines the evolution path to record the trajectory of $\bm m$. In fact, $\bm p_c$ is the weighted average of $\frac{\bm m^{(g+1)}-m^{(g)}}{\sigma}$ in every iteration:

\begin{equation}
\bm p_c^{(g+1)}=(1-c_c)\bm p_c^{(g)}+\sqrt{c_c(2-c_c)}\frac{\bm m^{(g+1)}-\bm m^{(g)}}{\sigma^{(g)}}.
\end{equation}

$\bm p_c^{(g)}\in\mathbb{R}^n$ is the evolution path at generation $g$ and $c_c\leq1$ the balance coefficient for historical accumulation and current generation.

The rank-1 update for the covariance matrix is expressed as

\begin{equation}
\bm C^{(g+1)}=(1-c_1)\bm C^{(g)}+c_1\bm p_c^{(g+1)}\bm p_c^{(g+1)^\mathsf T}.
\end{equation}

\subsection{Rank-$\mu$-Update}
The rank-$\mu$ update plays an important role when population size increases. It is actually the weighted maximum-likelihood estimation for the covariance matrix $\bm C$:

\begin{equation}
    \begin{split}
    \bm C^{(g+1)}&=(1-c_\mu)\bm C^{(g)}+c_\mu \frac{1}{\sigma^{(g)^2}}\\
    &=(1-c_\mu)\bm C^{(g)}+c_\mu \sum_{i=1}^{\mu}w_i\bm y_{i:\lambda}^{(g+1)} \bm {y}_{i:\lambda}^{(g+1)^{\mathsf T}}.
    \end{split}
\end{equation}

where $c_\mu$ is the learning rate for updating the covariance matrix, $w_{1,\ldots,\mu}\in \mathbb{R}$, $\sum_i w_i=1$ and $w_1\geq\ldots\geq w_\mu \geq 0$, $\bm y_{i:\lambda}^{(g+1)}=(\bm x_{i:\lambda}^{(g+1)}-\bm m^{(g)})/\sigma^{(g)}$

\subsection{Mean Vector Update}
The mean vector update is expressed as

\begin{equation}
\bm m^{(g+1)}=\bm m^{(g)}+c_m\sum_{i=1}^{\mu}w_i(\bm x_{i:\lambda}^{(g+1)}).
\end{equation}

where $c_m$ is the learning rate for updating the mean vector.

\begin{algorithm}
        \caption{CMA-ES}
    \begin{algorithmic}[1]

            \Require Optimization Objective (Fitness Function) $L()\in\mathbb{R}$, CMA-ES Hyper-parameter (Learning Step $\bm{\sigma}$, population size $\bm{\lambda}$, Elite-Population size $\bm\mu$, Initial Multivariate Normal Distribution (Covariance Matrix $\bm C$, Mean Vector $\bm m\in\mathbb{R}^{n}$))

            \Ensure Local Optimal Point $\bm p \in\mathbb{R}^n$

            \While{do not meet terminal conditions}
                \State Sampling Populations from current Multivariate Normal Distribution
                \State Calculate the Fitness Value for Population
                \State Select the Elite-Populations according to Fitness Value
                \State Recombine and update the Multivariate Normal Distribution (Rank-1 Update,Rank-$\mu$ Update, Mean Vector Update)

            \EndWhile
            \State
            \Return Local Optimal Point $\bm p \in\mathbb{R}^n$

    \end{algorithmic}
\end{algorithm}

We only introduce the above basic concept of CMA-ES in this paper considering that CMA-ES is a complicated subject. It is a necessary part of our black-box attack strategy as it can optimize without gradient information. Further, CMA-ES also involves the control of the learning-step size. The detailed analysis of CMA-ES can be found in \cite{b20}.

\section{Black-Box Attack Strategy}

Here, we introduce our black-box attack strategy.

To a classifier, the process of attack can be seen as the generation of an adversarial example. A black-box strategy means we must attack without any internal architecture information about the target model. The only accessible information about the target model is its output vector. The aim of attack is to find the perturbations that would lead to the misclassification of the target model. Furthermore, we can choose the attack target, i.e., the desired classification result after the input image is perturbed, and our strategy is able to manipulate the misclassification result on any class desired.

First, we must determine the attack target and the one-hot vector that could represent the attack target. Then, the formalization must be done to set the optimization problem of the perturbation.

The training step for the neural network can serve as a comparison for such an optimization problem.The supervised learning for the neural network will find the suitable network weight to minimize the loss function (like cross-entropy) at a given input and corresponding label as it is described in the following equation:

\begin{equation}\label{train}
\mathop{\arg\min_{\bm W}} \bm D(\bm Y,\bm F(\bm{X,W})),
\end{equation}

where $\bm X$ is the input image, $\bm W$ are the weights of the neural network, $\bm Y\in\mathbb{R}^n$ corresponds to the one hot label vector for input $\bm X$,
$\bm F()\in\mathbb{R}^n$ is the classifier function, i.e., network, $\bm D()\in\mathbb{R}$ is the loss function for supervised learning, e.g., cross-entropy.

For an adversarial attack, the network is trained, suggesting that $\bm W$ is fixed and the parameters that must be optimized are the perturbations $\bm \eta$. If we set the adversarial label $\bm Y'$, which represents the attack target for the corresponding input $\bm X$, the minimization process will finally find the perturbation that enables the input image $\bm X$ to fool the classifier:

\begin{equation}\label{final_optimize}
\mathop{\arg\min_{\bm{\eta}}} \bm D(\bm Y',\bm F(\bm X+\bm{\eta},\bm W))+\beta \parallel\bm \eta\parallel,
\end{equation}

where $\bm \eta\in\mathbb{R}^{w*h*c}$ is the noise of the original input image and $\parallel.\parallel$ is the norm of the tensor, e.g., $L_2, L_{\infty}$. To limit the perturbations, $\beta$ is the balance coefficient in multi-objective optimization.

We have thus set the optimization objective. The next step is to use the EA to solve this problem. Additionally, we need to set the terminal conditions for evolution iteration, as expressed in Algorithm 2.

\begin{algorithm}
        \caption{Evolution Attack}
    \begin{algorithmic}[1]

            \Require Trained Network $ \bm F()\in\mathbb{R}^n$,
            Input image $\bm{X}\in\mathbb{R}^{w*h*c}$,
             Attack Target $t\in\{0,1,\ldots,n-1\} $ \textbf{and} $t\not=$correct class,
             Fitness Function$D()\in \mathbb R$(like cross-entropy),
             CMA-ES Hyper-parameter (Learning Step $\bm{\sigma}$, population size $\bm{\lambda}$, Elite-Population size $\bm\mu$, Initial Multivariate Normal Distribution (Covariance Matrix $\bm C$, Mean Vector $\bm m\in\mathbb{R}^{n}$))

            \Ensure Adversarial Example (Perturbed Image)

            \State Using One Hot encoding to set the \textbf{Adversarial Target Vector} $\bm{Y'}\in\mathbb{R}^n$ based on Attack Target $t$

            \State Acquire the current classification result $cls=argmax\bm{F(X)}$
            \While{$cls\not=t$}

                \State Sampling Perturbation $\bm \eta\in\mathbb{R}^{w*h*c}$ from the current Multivariate Normal Distribution as the population $\bm{P}=\{\bm\eta_0,\bm\eta_1,¡­,\bm\eta_{\lambda-1}\}$
                \State Predict the Perturbed Images $\hat{\bm Y_i}=\bm F(\bm{X}+\bm{\eta}_i)(i\in\{0,1,\ldots,\lambda-1\})$ and calculate the Fitness Values $D(\hat{\bm Y_i},\bm{Y'})$ respectively.
                \State Select the Elite-Populations and Recombination to update the Multivariate Normal Distribution

                \State $cls=argmax\bm {F(X+\eta_0)}$ ($\bm\eta_0$ is the best perturbation base on Fitness Value)

            \EndWhile
            \State
            \Return Adversarial Example=($\bm X+\bm\eta_0$)
    \end{algorithmic}
\end{algorithm}

\section{Experiment}

To test our black-box attack strategy, we require a trained network, and we therefore trained two classifiers that meet the baseline criteria: 1) LeNet on MNIST, which can achieve 98.9\% accuracy on the test set, and 2) simple ResNet$-$18 on Cifar$-$10, which can achieve 90.1\% accuracy on the test set. The architectures of these networks are described in Tables~\ref{LeNet} and ~\ref{ResNet}, respectively.

\begin{table}[htbp]
\caption{LeNet Architecture}
\begin{center}
\begin{tabular}{cccc}
\hline
\textbf{Layer Type} &\textbf{Parameter}&\textbf{Padding}&\textbf{Activation} \\
\hline
Conv & Kernel:3x3\_20, Stride=1 & SAME& ReLu  \\
MaxPool & 2x2 &&  \\
Conv & Kernel:3x3\_20, Stride=1 & SAME& ReLu   \\
MaxPool & 2x2&  \\
Flatten\\
FC&500 (Number of neurons)&&ReLu\\
FC&10 (Number of neurons)&&SoftMax\\
\hline
\end{tabular}
\label{LeNet}
\end{center}
\end{table}

\begin{table}[htbp]
\caption{ResNet-18 Architecture}
\begin{center}
\begin{tabular}{cccc}
\hline
\textbf{Layer Type} &\textbf{Parameter}&\textbf{Padding}&\textbf{Activation} \\
\hline
Conv & Kernel:3x3\_64, Stride=1 & SAME& ReLu  \\
Residual Block &Kernel:3x3\_64, Stride=1&SAME&ReLu \\
Residual Block &Kernel:3x3\_128, Stride=1&SAME&ReLu \\
MaxPool & 2x2 &&  \\
Residual Block &Kernel:3x3\_256, Stride=2&SAME&ReLu \\
Residual Block &Kernel:3x3\_512, Stride=1&SAME&ReLu \\
MaxPool & 2x2&  \\
Conv & Kernel:3x3\_512, Stride=1 & SAME& ReLu   \\
Flatten\\
FC&10 (Number of neurons)&&SoftMax\\
\hline
\end{tabular}
\label{ResNet}
\end{center}
\end{table}

In the MNIST experiment, we tested different EAs, including a simple GA, PEPG, OpenAI ES, and CMA-ES. We also tested different population sizes, i.e., 25 and 100. The elite populations ratio was set to 50\%, initial mean vector to $\bm 0^{w*h*c}$, and learning step size to 0.05. For an adversarial target, we chose the second-most likely class predicted by the classifier as the original input image. In this case, the evolution iteration would keep attacking until the desired misclassification occurred. For regularization in \eqref{final_optimize}, we set $\beta$ as 1.

These terminal conditions ensure that our attack method could generate an adversarial example in 100\% of the cases. In other words, the quality of the adversarial example is the most important issue for evaluating the strategy. We define the quality of the adversarial example as the similarity between the original image and adversarial example. We therefore measured the ${L_1 Norm}$, ${L_2 Norm}$, and $L_{\infty} Norm$ of perturbations. Additionally, we used the \textbf{structural similarity index} (SSIM) to evaluate the two images' similarity.

As a realization of the structural similarity theory, the SSIM defines the structural information from the perspective of image composition independent of brightness and contrast, reflecting the properties of the object structure in the scene. The mean is used as the estimate of the brightness, the standard deviation as the estimate of the contrast, and the covariance as a measure of the degree of structural similarity. To a certain extent, the SSIM can reflect the difference between the original and perturbed images in human visual perception. SSIM values range from 0 to 1, where 1 means that two images are exactly the same.
\begin{table}[htbp]
\caption{MNIST Experiment: Performance of Different EAs in Adversarial Attack}
\begin{center}
\begin{tabular}{|c|c|c|c|c|}
\hline
 &\multicolumn{4}{|c|}{\textbf{Similarity Evaluation}} \\
\cline{1-5}
\textbf{EA} & ${L_1 Norm}$&${L_2 Norm}$ &$L_\infty Norm$&\textbf{\textit{SSIM}} \\
\hline
GA$_{(25)}$ & 34.5883 & 1.9533 &0.3006 &0.8037 \\
\hline
GA$_{(100)}$ & 32.0961 &1.7931 &0.2671 &0.8151 \\
\hline
PEPG$_{(25)}$ & 38.0039 &2.1237 &0.2875 &0.7914 \\
\hline
PEPG$_{(100)}$ & 32.4222 &1.7693 &0.2122 &0.8164 \\
\hline
OpenAI$_{(25)}$ & 35.1598 & 1.9473 &0.2523&0.8034 \\
\hline
OpenAI$_{(100)}$ & 30.8387 &1.6566 &\textbf{0.1841} &0.8251 \\
\hline
CMA-ES$_{(25)}$ & \textbf{25.4210} &\textbf{1.4649} &0.2556 &\textbf{0.8470} \\
\hline
CMA-ES$_{(100)}$ & 32.2897 & 1.8284 &0.3024 &0.8117 \\
\hline
\end{tabular}
\label{tab3}
\end{center}
\end{table}

From Table.~\ref{tab3}, it can be seen that a simple GA, PEPG, and OpenAI-ES improve their optimization effect with increased population size. However, CMA-ES performed better in population sizes of 25 than in population sizes of 100 without any regularization. We think the reason for this phenomenon is the recombination method in the algorithm. For simple optimization problems, the increase of population size leads to an increase in the number of elite populations, which brings some noise to the distribution update. Since images in MNIST have 28*28*1 pixels and images in Cifar$-$10 have 32*32*3 pixels, it is not suitable to compare the $L_1 Norm$ and $L_2 Norm$ of the different datasets. We only observe the performance of different population sizes in the same dataset. In Table.~\ref{tab5}, for the Cifar$-$10 experiment, it can be seen that, for large population sizes, CMA-ES performs better. Owing to the larger number of dimensions in optimization, the opposite result is shown for the MNIST experiment. Therefore, one cannot just say which population size is better, but performance is determined by the actual situation. For low-dimensional problems, large population size does not guarantee better performance. For high-dimensional problems, the optimization result may be improved because of more populations. The quantitative relationship between dimension and population size merits more attention. Further, another factor is the weighted method of elite populations. A smoother weighted method means more global searching, in other words, the approach may not be efficient enough. In our attack method, the more efficient optimization method means a higher-quality adversarial image. Thus, more time is still needed to explore how to improve the optimization method in such situations.

\begin{table}[tbp]
\caption{MNIST Experiment: Performance of Different Regularizations on Adversarial Attack}
\begin{center}
\begin{tabular}{|c|c|c|c|c|}
\hline
 &\multicolumn{4}{|c|}{\textbf{Similarity Evaluation}} \\
\cline{1-5}
\textbf{EA} & ${L_1 Norm}$&${L_2 Norm}$ &$L_\infty Norm$&\textbf{\textit{SSIM}}  \\
\hline
CMA-ES$_{(25)}$ & {25.4210} &{1.4649} &0.2556 &{0.8470} \\
\hline
CMA-ES$_{(25)-L_2}$ & \textbf{24.2534} &\textbf{1.4062} &0.2486 &\textbf{0.8531} \\
\hline
CMA-ES$_{(25)-L_\infty}$ & {25.5840} &{1.4610} &\textbf{0.2159} &0.8497 \\
\hline
CMA-ES$_{(100)}$ & {32.2897} &{1.8284} &0.3024 &{0.8117} \\
\hline
CMA-ES$_{(100)-L_2}$ & {31.4301} &{1.7800} &0.2964 &{0.8153} \\
\hline
CMA-ES$_{(100)-L_\infty}$ & {25.8740} &{1.4766} &0.2315 &{0.8484} \\
\hline
\end{tabular}
\label{tab4}
\end{center}
\end{table}

\begin{table}[tbp]
\caption{Cifar-10 Experiment: Different Regularizations on Adversarial Attack}
\begin{center}
\begin{tabular}{|c|c|c|c|c|}
\hline
 &\multicolumn{4}{|c|}{\textbf{Similarity Evaluation}} \\
\cline{1-5}
\textbf{EA} & ${L_1 Norm}$&${L_2 Norm}$ &$L_\infty Norm$&\textbf{\textit{SSIM}}  \\
\hline
CMA-ES$_{(25)}$ & {150.8968} &3.4296 &0.2329 &{0.7908} \\
\hline
CMA-ES$_{(25)-L_2}$ &150.2518 &3.4152 &0.2313 &{0.7921} \\
\hline
CMA-ES$_{(25)-L_\infty}$ & 150.6225 &3.4229 &{0.2295} &0.7915 \\
\hline
CMA-ES$_{(100)}$ & 124.5613 & 2.8314 & 0.1925 &{0.8370} \\
\hline
CMA-ES$_{(100)-L_2}$ & \textbf{123.9939} &\textbf{2.8182} &0.1908 &{0.8384} \\
\hline
CMA-ES$_{(100)-L_\infty}$ & 124.5772 &2.8310 &\textbf{0.1884} &\textbf{0.8474} \\
\hline
\end{tabular}
\label{tab5}
\end{center}
\end{table}

Another issue worth discussing is regularization. From Table.~\ref{tab4}, it can be seen that additional regularization indeed improves the performances of our attack method. The optimization difficulty increases with added regularization, and turns out to be the multi-objective optimization problem. In such situations, we think a large-populations-size CMA-ES could work better. Experimental results show that a large-population-size CMA-ES exhibits more obviously improved performance, while a small-population-size CMA-ES still shows better problem-solving ability.
In Table.~\ref{tab5}, for the Cifar$-$10 experiment, the benefits from different regularizations are not obvious enough to say which regularization is better. We regard the SSIM as the most important index in our evaluation. Our thinking is, if large population size can bring a more direct searching direction in the multi-objective optimization problem, then the $L_\infty Norm$ may be the better choice to limit perturbations, because in both the MNIST and Cifar$-$10 experiments,  $L_\infty Norm$ helps more than $L_2 Norm$ in large-population-size CMA-ES. However, we are not very confident in this conclusion, and more detailed experiments should be conducted to vet this idea.

The hyper-parameter of CMA-ES, as the most indispensable and important part of our evolution attack method, will directly influence the quality of an adversarial attack. As with the evolution algorithm, it is impossible to say which hyper-parameter is definitely the best. However, our experiment fortunately shows that it is perfectly suitable for our evolution attack.

\section{Acknowledgements}
This work is supported by the Project supported by the National Key Research and Development Program (Grant no. 2016YFB0100901), the Shanghai Municipal Commission of Economy and Informatization (Grant no. 2018RGZN02050), and the Science and Technology Commission of Shanghai (Grant no. 17DZ1100202; 16DZ1100700).

\vspace{12pt}
\color{red}

\end{document}